\documentclass[letterpaper]{article}
\usepackage{aaai2027}
\usepackage[hyphens]{url}
\usepackage{graphicx}
\usepackage{natbib}
\usepackage{caption}
\DeclareCaptionStyle{ruled}{labelfont=normalfont,labelsep=colon,strut=off}
\usepackage{multirow}
\usepackage{array}
\usepackage{booktabs}
\usepackage{amsmath}
\usepackage{amssymb}
\nocopyright

\title{Getting the Parameters Right: A Difficulty-Graded Benchmark and Probe-Guided Training for LLM Tool Calls}

\author{
    Guoyao Yu,
    Xiaoqing Sun,
    Ziqi Huang,
    Shaojing Fan,
    Zhongyi Zhang,
    Xiaomeng Hu,
    Xiaobo Xue,
    Yangyang Shi,
    Xiong Xiao,
    Yang Song,
    Biao Lyu,
    Rong Wen,
    Xing Li,
    Qinming He,
    Shunming Zhu,
    Zhenguang Liu\corresponding
}
\affiliations{
}

\begin{document}
\maketitle

\begin{abstract}
Large language model agents derive much of their capability from tool
use. Existing research on tool use has largely focused on selecting
the right tool and orchestrating the order of calls. However,
correctly filling the parameters of a tool call is equally critical
for successful execution and has received far less attention. In
domains such as cloud networking, even frontier models correctly
complete fewer than half of tool calls. Inspired by recent analyses
showing that LLM hidden states encode rich information about model
predictions, we discover that while the model generates a parameter value,
its hidden state contains a strong correctness signal: a simple
linear probe can accurately predict whether the value will be
correct. Based on this observation, we propose a unified probe-guided
framework with two complementary approaches: probe-filtered
bootstrapped training (PBT), which uses the probe to filter reliable
self-generated calls for fine-tuning, and probe-guided reranking
(PGR), which uses the probe to select better candidates during
inference. To support systematic evaluation, we release
\textsc{ParamBench}, a benchmark built from real cloud-network APIs
that categorizes every instance into five difficulty levels according
to parameter nesting depth, cross-parameter dependencies, and the
reasoning required to derive values from earlier calls. Extensive
experiments across 5 open models on \textsc{ParamBench} and 6
external benchmarks demonstrate that our method substantially
improves parameter generation, raising the average exact match from
19.7\% to 59.6\%.
\end{abstract}

\section{Introduction}
\label{sec:intro}

Tool use is now the core capability of large language model (LLM)
agents. It is what connects an agent to the world outside its context
window: the agent perceives its environment by reading what its API
calls return, and acts by issuing further calls. Making tool use work
well is not simple. Industry has built invocation standards such as
the Model Context Protocol~\citep{mcp-anthropic24} and function
calling~\citep{openai-funccall23}; research has studied when to call
a tool, which APIs to call, and in what order to call
them~\citep{toolllm-iclr24,gorilla-neurips24,apibank-emnlp23,toollearning-csur25}.
However, in many real-world scenarios that involve complex tool
calls, getting the parameters of a call right is itself a hard
problem, and it has received far less systematic study.

\begin{figure}[t]
\centering
\includegraphics[width=\columnwidth]{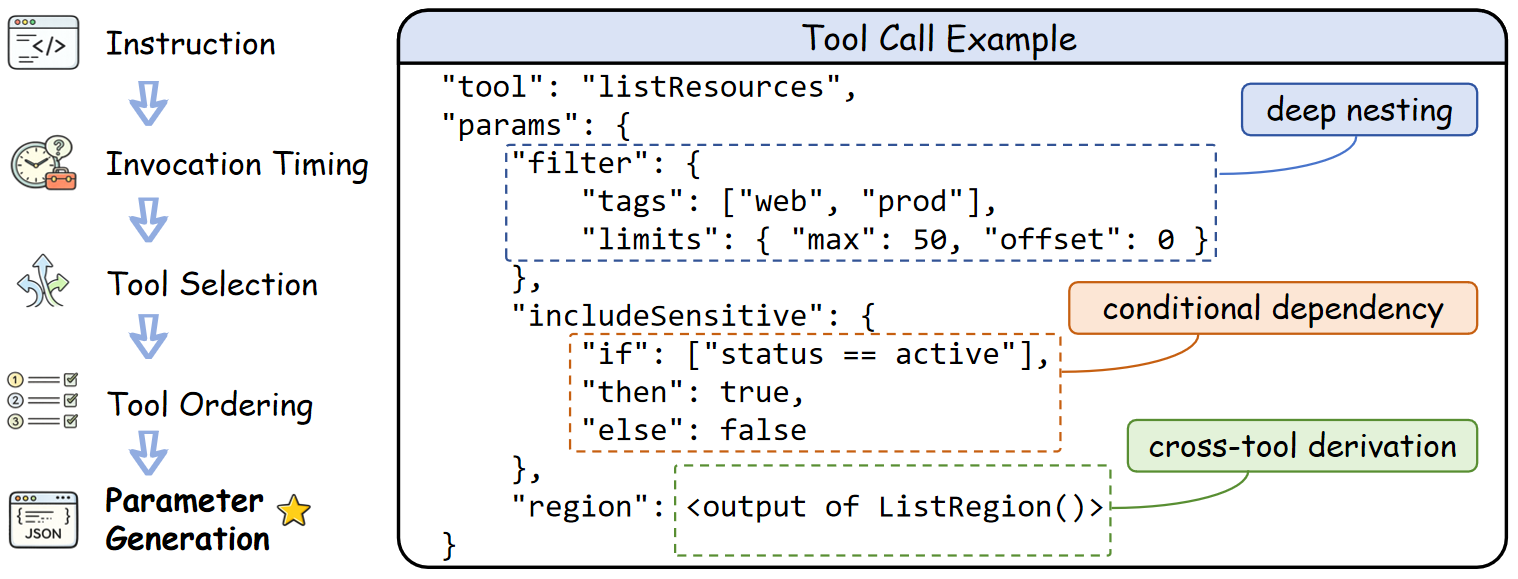}
\caption{Tool Use Overview and Parameter Generation}
\label{fig:teaser}
\end{figure}

Formally, we define this step as \emph{parameter generation}: given
an instruction, the API schemas, and the results of earlier calls,
the model must fill every parameter of a tool call with a correct
value. There are three structural properties that make it hard:
\textbf{(i) deep nesting}: parameters can be objects nested several
layers deep; \textbf{(ii) conditional dependency}: whether a field is
required or valid can depend on other fields; \textbf{(iii)
cross-call derivation}: some values must be derived from the results
of earlier calls. Figure~\ref{fig:teaser} illustrates the task
and the three difficulties. In our tests, Claude Opus 4.7 fills only 33.4\% of
the calls correctly on NESTFUL in the 0-shot setting.

Inspired by prior work showing that a model's hidden states carry
signals such as whether its own output is true~\citep{saplma-emnlp23,geomtruth-colm24,haloscope-neurips24},
we examine a similar signal at the parameter level and find that a
linear probe on the hidden state predicts whether a parameter value
will be correct, with an in-domain AUC of 0.986.
Building on this observation, we use the probe in two places: probe-filtered
bootstrapped training (PBT) on the training side and probe-guided
reranking (PGR) on the inference side. PBT targets the common case
where labeled answers are scarce and unlabeled instructions are
plentiful. A seed model is first fine-tuned on the small labeled set, and
then generates answers for the unlabeled instructions; the answers
that pass the probe are added to the labeled set, and the base model
is fine-tuned on the enlarged set to produce the final model. In PGR, the model samples several
candidate calls at inference time, the probe scores them, and a
selection strategy (candidate-level, field-level, or field-set)
picks the final call. The two sides pay different costs: PBT changes the model once at
training time, while PGR leaves the model unchanged and spends
extra samples at inference time; either can be used alone, and they
can also be combined.

For evaluating parameter generation, existing benchmarks match the
generated call against a reference
call~\citep{bfcl-icml25,apibank-emnlp23,sealtools-nlpcc24,nestful-emnlp25}.
They do not classify where parameter values come from, and they do
not grade how hard an instance is.
We therefore present \textsc{ParamBench}, built from real agent
tool-call execution traces in cloud networking. Based on nesting
depth, cross-parameter dependency, and the reasoning required to
derive values from earlier calls, we design a fixed rule that grades
the parameter generation difficulty of every instance into five
levels, L1 to L5. This gives a fine-grained view of how well a model
fills tool-call parameters.

We evaluate PBT and PGR on \textsc{ParamBench} and 6 external
tool-use benchmarks, including BFCL and API-Bank, over 5 open models
such as Qwen3-8B and Gemma-4-12B. Averaged over models and datasets,
plain supervised fine-tuning raises exact match from 19.7\% to
51.6\%, and PBT raises it to 59.6\%. On the two deepest-nesting
benchmarks, PGR adds a further 4.6 points at inference time.
Moreover, compared with frontier models such as Claude Opus 4.7 and
GPT-5.4, a Qwen3-8B model with PBT and PGR reaches the frontier
level on all 7 datasets and is the best on 3 of them.

Our main contributions are as follows.
\begin{itemize}\itemsep=0pt
\item[\textbf{C1}] \textbf{(New focus).} We put parameter generation
at the center of tool-use research, define it formally, and analyze
its three difficulties: deep nesting, conditional dependency, and
cross-call derivation.
\item[\textbf{C2}] \textbf{(Internal-signal supervision).} We
demonstrate that hidden-state correctness signals can provide
effective supervision for structured tool-call parameter generation.
Building on this observation, we propose probe-filtered bootstrapped
training (PBT) for selecting reliable pseudo-labels during
self-training, and probe-guided reranking (PGR) for selecting the
best candidate at inference.
\item[\textbf{C3}] \textbf{(ParamBench Dataset).} For fine-grained
evaluation of parameter generation, we release \textsc{ParamBench}, a
benchmark built from real cloud-network agent traces, and grade each
tool-call instance into five difficulty levels by the structural
properties of its parameters.
\item[\textbf{C4}] \textbf{(Extensive evaluation).} We run
large-scale comparisons and ablations over 7 benchmarks and 5 base
models, compare 4 selection signals and 4 leading tool-use models,
and show that PBT and PGR are the most effective strategies.
\end{itemize}

\section{Related Work}
\label{sec:related}

\paragraph{Tool learning and function calling.}
Equipping language models with external tools has grown into a mature
research area, commonly termed \emph{tool
learning}~\citep{toollearning-csur25,toolsurvey-fcs25}, supported by
widely adopted standards such as the Model Context
Protocol~\citep{mcp-anthropic24} and OpenAI's function-calling
interface~\citep{openai-funccall23}. Research in this area has
centered on three questions. The first is \emph{when} to call a tool:
Toolformer learns where an API call should be inserted in the
text~\citep{toolformer-neurips23}, and MetaTool benchmarks the
decision of whether to invoke a tool~\citep{metatool-iclr24}. The
second is \emph{which} tool to select: ToolLLM retrieves relevant
APIs from a pool of more than 16{,}000
candidates~\citep{toolllm-iclr24}, Gorilla couples retrieval with
instruction tuning~\citep{gorilla-neurips24}, and Re-Invoke rewrites
the query for zero-shot retrieval~\citep{reinvoke-emnlp24}. The third
is in \emph{what order} to compose several calls: ReAct alternates
reasoning steps and tool actions~\citep{react-iclr23}, and
LLMCompiler plans a graph of calls that can run in
parallel~\citep{llmcompiler-icml24}. These works all treat
one accurate call as a single unit, and parameter generation
inside the call has not been studied as a separate problem.

\paragraph{Function-calling methods.}
In all current research, the parameter values that fill each call are
left to free-form generation. Constrained decoding guarantees that
the output parses, but it cannot tell which legal value is the right
one~\citep{outlines-arxiv23,xgrammar-mlsys25}. Tool-call SFT, on
benchmark or large synthetic call data, covers the hard cases
only where the training data happens to contain
them~\citep{toolace-iclr25,toolllm-iclr24,gorilla-neurips24,xlam-naacl25}.
Few-shot prompting covers only what the examples show.
ReAct-style reasoning gets the value right only when the reasoning
happens to reach it~\citep{react-iclr23}. To date, a method that specifically addresses the difficulties of
parameter generation is still missing.

\paragraph{Tool-use benchmarks.}
On the evaluation side, existing benchmarks center on tool selection
and multi-step orchestration~\citep{apibank-emnlp23,nestful-emnlp25}.
Where they do inspect arguments, the measure is coarse: BFCL folds
argument correctness into a single call-success
score~\citep{bfcl-icml25}, $\tau$-bench reveals wrong parameters
only through the final task outcome~\citep{taubench-iclr25}, and
Seal-Tools reports only an overall parameter match
rate~\citep{sealtools-nlpcc24}. NesTools is the closest prior effort
to score nested-parameter filling as a
separate axis, and current models struggle on
it~\citep{nestools-coling25}. Even so, no existing dataset scores parameter generation at the
field level and grades every instance by difficulty.

\paragraph{Internal signals of model correctness.}
A line of work shows that a model's hidden states encode whether its
own output is correct, often more reliably than its stated confidence
or its token probabilities. Early probing studies established this on
factual statements: a simple probe on hidden activations tells
whether a statement is true, and truth even appears as a linear
direction in the representation
space~\citep{saplma-emnlp23,geomtruth-colm24}. Hidden states,
embeddings, and gradients of a completed answer likewise detect
hallucinated text~\citep{egh-emnlp24,llmsknowmore-iclr25}, and
follow-up work reads the signal before or during generation, so an
error can be predicted before the answer is
complete~\citep{internalstatesrisk-blackboxnlp24,seps-arxiv24,earlydetecthallu-kdd24}.
The signal can also be acted on: likely errors can be
flagged~\citep{haloscope-neurips24}, or generation can be guided
toward a more truthful answer~\citep{iti-neurips23}. In tool use,
however, this signal is nearly untouched: the only attempt detects a
bad call after it is fully generated, with one yes-or-no label for
the whole call~\citep{toolselhalluprobe-arxiv26}.

\section{Problem Formalization}
\label{sec:problem}

In production tool-use systems, selecting the appropriate API can
often be addressed through retrieval or routing, whereas correctly
instantiating structured API parameters remains a major challenge.
This section formalizes tool-call parameter generation as a
value-filling task and identifies its three structural sources of
difficulty: deep nesting, inter-field conditional dependencies, and
cross-call value derivation.

Given a target API schema $\mathcal{S}$ (each
parameter's name, type, required flag, nesting structure, and field
description), a natural-language instruction
$\mathcal{I}$, and a data-flow context $\mathcal{D} = \{(t_i, o_i)\}$
of upstream tool calls and their outputs, the task is to produce a
parameter instance $\mathbf{p}$ that is format-compliant under
$\mathcal{S}$ and value-correct against a reference
$\mathbf{p}^{\ast}$. In practice, only value correctness is hard: a preliminary audit of 7
frontier models shows that only 2.1\% of their failed calls
violate the schema, while the others are schema-valid but
value-wrong. The three challenges below are the main reasons the values
go wrong.

\paragraph{CH-1: Deep Nesting.}
Enterprise API parameters are often not flat key-value maps but
nested objects and object lists. A typical example is the
\texttt{Filters} parameter of \texttt{ListTransitRouterRouteEntries}:
{\footnotesize
\begin{verbatim}
Filters = [ {Key: DestinationCidrBlock,
             Value: [10.0.0.0/16]},
            {Key: Status, Value: [Active]},
            ... ]
\end{verbatim}
}
a list of filter objects whose \texttt{Value} field is itself a list:
an array inside an object inside an array. A constrained decoder can keep this
shape well-formed, but the right literal must still be placed at the
right depth.

\paragraph{CH-2: Inter-Field Conditional Dependencies.}
Some parameters are required, or take constrained values, only when a
sibling parameter takes a specific value: \texttt{NextHopId} is
required when \texttt{NextHopType="RouterInterface"}; a
\texttt{peerInfo} sub-object is required only when \texttt{type="VBR"}.
The same schema therefore admits several valid shapes, selected by one
controlling field. In the public schemas of production APIs these
rules are rarely written in machine-readable form; they live in
human-readable field descriptions, and the model must infer which shape the current
situation calls for.

\paragraph{CH-3: Cross-Call Value Derivation.}
Many correct values are not in the instruction but in an earlier
tool's output: the \texttt{VpcId} returned by \texttt{DescribeVpcs}
is an input to \texttt{DescribeVSwitches}. The model must find the
right field in an upstream response that is often dozens of fields
wide and several layers deep, extract the value, and place it into
the right downstream slot. Picking the wrong field, or the wrong
element of a list, is a frequent failure.

\section{Probe-Guided Training and Reranking}
\label{sec:method}

\begin{figure*}[t]
\centering
\includegraphics[width=0.98\textwidth]{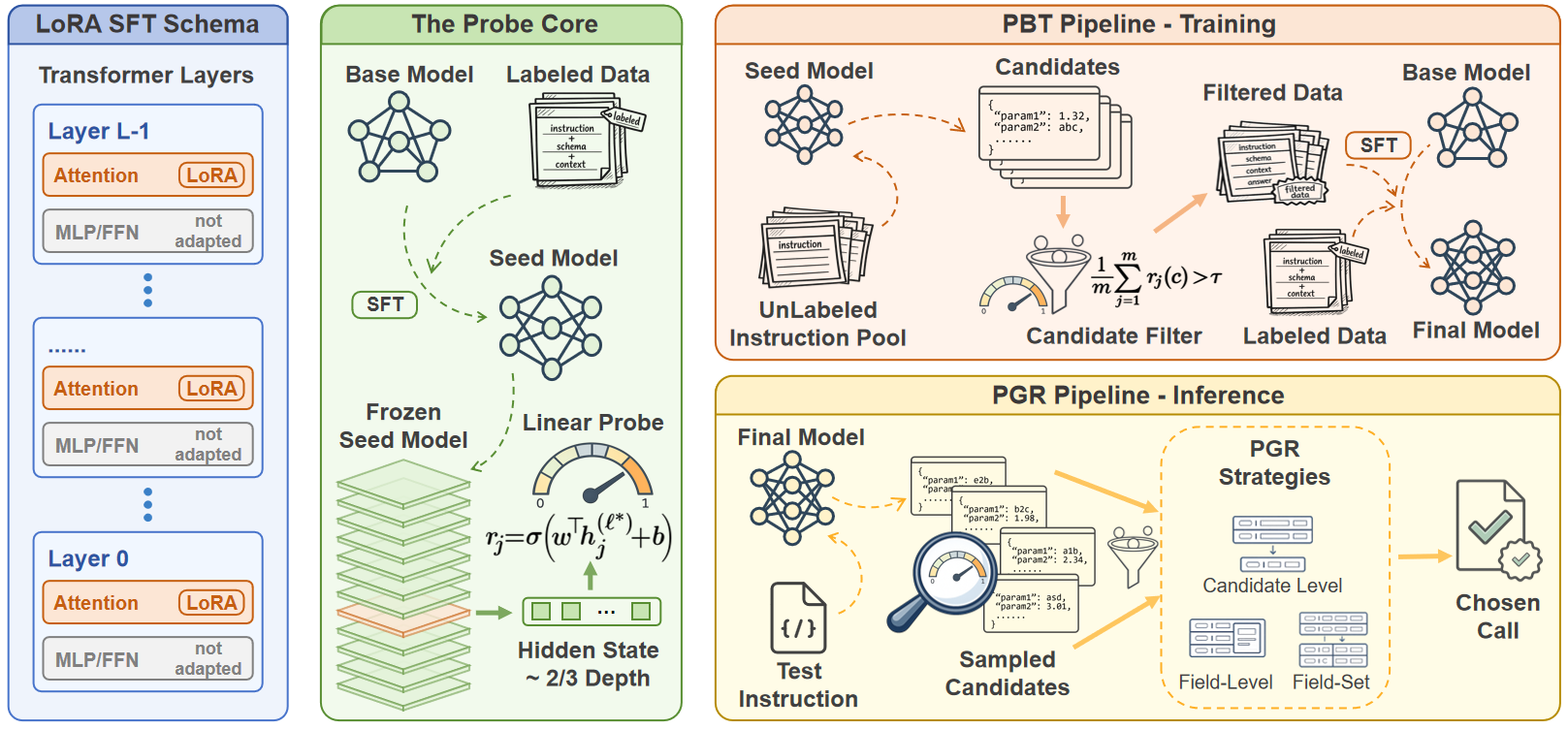}
\caption{Overview of our framework. Left: the base model is
fine-tuned with LoRA, which adapts only the attention layers. Middle: a
linear probe is trained on labeled data, using hidden states from
roughly two-thirds of the seed model's depth. Top right: the trained
probe defines a candidate filter that converts unlabeled data into
pseudo-labeled data, enabling Probe-Filtered Bootstrapped Training
(PBT). Bottom right: at inference time, the same probe reranks
sampled candidates to select the final call, yielding Probe-Guided
Reranking (PGR).}
\label{fig:method}
\end{figure*}

Inspired by a line of work that reads correctness signals from a
model's hidden
states~\citep{saplma-emnlp23,geomtruth-colm24,haloscope-neurips24}, we
build a linear probe: it reads the hidden state just before the model
writes a parameter value, and predicts whether that value will be
correct. We then use this signal in
two places. On the training side, probe-filtered bootstrapped training
(PBT) uses it to produce more useful supervised fine-tuning (SFT) data
for the model. On the inference side, probe-guided reranking (PGR) uses
it to rerank sampled parameter candidates. Figure~\ref{fig:method}
gives an overview.

\subsection{The Probe: Correctness Before Emission}
\label{subsec:observation}

To obtain a reliable probe signal, we keep the model frozen throughout.
At the position just before the model writes a parameter value, we take
the hidden state and feed it to a logistic regression classifier, which
outputs the probability that the value will be correct. We train one
probe per layer and keep the layer that performs best, which sits at
roughly two-thirds of the network depth
(Figure~\ref{fig:auc} in the appendix).

Formally, let $t_j$ be the decision point of the $j$-th parameter, which is the
position just before the model writes its value, and let
$h_j^{(\ell)}$ be the hidden state that layer $\ell$ produces at $t_j$.
The probe is a logistic regression over this vector:
\begin{equation}
r_j \;=\; \sigma\!\left(w^\top h_j^{(\ell^*)} + b\right),
\label{eq:probe}
\end{equation}
where $w$ and $b$ are the learned weight vector and bias, $\sigma$ is
the sigmoid function that turns the score into a probability, $\ell^*$
is the best layer chosen as above, and $r_j$ is the predicted
probability that parameter $j$ will be correct. To train $w$ and $b$,
we take training-side instructions that have gold answers, let the
model sample candidate values, and label each parameter $y_j = 1$ if
its value matches the gold answer and $y_j = 0$ otherwise; $w$ and
$b$ are then fit by minimizing the standard binary cross-entropy loss
between $r_j$ and $y_j$.

A probe built this way reaches an in-domain AUC of 0.986, well above
the baseline of token log-probability (0.914), a common signal for
the confidence of model outputs~\citep{pik-arxiv22}.
The signal is also robust: the probe signal survives the move to
the next training checkpoint (AUC 0.982), and across model sizes, before and
after fine-tuning, the AUC stays between 0.93 and 0.99, always above
log-probability. One property to note is that the probe is
specialized: it is trained for a given dataset, model, and sampling
temperature, and its accuracy degrades outside that
setting. In our design, each domain therefore trains its own probe on
its own training set, which keeps the probe in the setting it
handles best.

\subsection{Probe-Filtered Bootstrapped Training}
\label{subsec:pbt}

In real tool-use applications, instructions and gold answers come at
very different costs. Instructions are nearly free, since user
requests accumulate in production logs on their own. Gold answers
are expensive, because each one must be written and checked, field by
field, by an expert who knows the API. Most domains therefore end up
with a small labeled set, denoted $\mathcal{L}$, and a large pool of
unlabeled instructions, denoted $\mathcal{U}$.

The natural way to use $\mathcal{U}$ is self-training: let the model
answer the unlabeled instructions and train on its own answers. The
difficulty lies in the fact that many of these answers are wrong, and they are hard
to catch: there is no gold answer to compare against, and schema
checks pass nearly all wrong calls. The probe fills this gap: it predicts correctness without seeing
the gold answer.
\emph{Probe-filtered bootstrapped training} (PBT) builds the judge
into the self-training loop and runs in the following five steps.
\begin{enumerate}
\item Fine-tune the model on $\mathcal{L}$ with
LoRA~\citep{lora-iclr22}. The result is the \emph{seed model}.
\item Build the probe of Equation~\eqref{eq:probe} for the seed
model, on $\mathcal{L}$ only: the seed model samples candidate values,
each labeled against the gold answer, and $w$ and $b$ are
fit.
\item Run the seed model over $\mathcal{U}$ and collect several
candidate calls for each instruction.
\item Score every candidate with the probe. The score of a candidate
call $c$ is the mean probe score over the $m$ parameters it fills:
\begin{equation}
s(c) \;=\; \frac{1}{m}\sum_{j=1}^{m} r_j .
\label{eq:pbt-score}
\end{equation}
For each instruction, keep the highest-scoring candidate only if its
score passes a threshold $\tau$. Merge the kept instruction--call
pairs with $\mathcal{L}$; call the result $\mathcal{L}^{+}$.
\item Fine-tune the base model on $\mathcal{L}^{+}$ with LoRA. The
result is the final model.
\end{enumerate}
The threshold $\tau$ is set per domain on the training set, by
sweeping a small grid and keeping the value that
yields the best filtered training set; typical values are 0.9 and
0.95.

\subsection{Probe-Guided Reranking}
\label{subsec:rerank}

At inference time, the final model generates a pool of candidate calls
for each instruction: one greedy decode and several sampled ones. The
probe scores every candidate, the greedy one included, with the same
score $s(c)$ as in Equation~\eqref{eq:pbt-score}.

A decision strategy turns these scores into a single call to submit.
Our strategies form 3 families that differ in the unit over which
they decide.
\emph{Candidate-level} strategies return one whole candidate and
differ only in the ranking score: the probe score $s(c)$, the
candidate log-probability, or a weighted sum of the two. 
\emph{Field-level} strategies take
the set of fields from one candidate and then choose each value on its
own, from whichever candidate scores best on that field.
\emph{Field-set} strategies do the reverse: they keep the greedy
values and decide only which fields survive, preferring the smallest
field set that the probe still rates highly. A field is dropped when
its own probe score is low, when few candidates emit it, or when a
candidate that scores nearly as well omits it.

Each family also has a version that stays close to the greedy decode
and changes it only on clear evidence. We do not fix one strategy in
advance. For each domain the training set is split into 5 parts, the
strategies are compared on 4 of them, and the winner is applied to the
fifth. The cost of PGR is
small: the probe is shared with PBT, and scoring one candidate adds
one forward pass.

\section{ParamBench}
\label{sec:parambench}

In this section, we present \textsc{ParamBench}, a benchmark for
tool-call parameter generation. As discussed in
Section~\ref{sec:related}, existing benchmarks check a tool call only
as a whole and do not grade how hard its parameters are to fill. 
In contrast, \textsc{ParamBench} shifts the focus to the
parameter values: \textit{once the right tool is chosen, can the model
correctly fill the parameters that the call requires?} The task
follows the value-filling setup of Section~\ref{sec:problem}.
\textsc{ParamBench} draws a large pool of tool-call instances from
real agent traces and the schemas of 81 cloud-network APIs, filters
them with structural checks, and grades every instance into five
difficulty levels by the structural features of its parameters,
giving 1{,}022 instances in total. Each instance keeps the standard
JSON-Schema call format and can be directly loaded and used by
existing function-calling evaluation tools.

\subsection{Deterministic Five-Level Difficulty Scale}
\label{subsec:parambench-grading}

Each \textsc{ParamBench} instance pairs a target API schema with a
natural-language instruction, an upstream context with the outputs
of earlier calls, and a gold answer. From these parts, four
complexity scores are computed automatically:
\begin{itemize}\itemsep=0pt
\item \texttt{nesting\_depth} (\texttt{d}): the deepest JSON nesting
among the required input fields;
\item \texttt{num\_conditional\_dependencies} (\texttt{c}): how many fields
depend on the value of a sibling field;
\item \texttt{num\_upstream\_transfers} (\texttt{t}): how many inputs
come from a prior tool's output;
\item \texttt{upstream\_extract\_depth} (\texttt{e}): how deep the
transferred value sits in that output.
\end{itemize}

A fixed rule (Table~\ref{tab:bench-distribution}, left) turns the
four scores into a level: it checks the levels in order and assigns
the first one that fits, so the level rises as nesting gets deeper,
conditional dependencies appear, and more values must be derived
from upstream outputs. Because the rule is a fixed function, every
instance gets its level the same way, with no human judgment.

\subsection{Dual-Source Construction}
\label{subsec:parambench-construction}

\textsc{ParamBench} instances come from two sources: human-verified
multi-step diagnosis traces from an industrial cloud-network agent system, 
and synthesis directly from the 81 frozen API schemas.

\paragraph{Trace extraction.} Each trace is a sequence of verified
tool calls, in which every step records the tool it called, the
parameters it used, and the output it returned. A converter walks
each trace and turns every executable step with a non-trivial
parameter into one instance, which consists of an instruction, an
upstream context, and a gold answer. The instruction is assembled
from the trace's task description and the step's context; the
upstream context collects the outputs of the preceding steps,
trimmed to the fields the call actually reads; and the gold answer
is the step's verified parameters. Difficulty rises naturally with a
step's position, because a first step needs no upstream value while
a later step must copy several from earlier outputs. Before release,
every instance is checked to confirm that the gold answer validates
against the frozen schema, and resource ids are replaced by
anonymized ids in the same format.

\paragraph{Schema-driven synthesis.} The hard instances are
synthesized directly from the 81 frozen API schemas. Because each
schema's complexity profile sets the difficulty level that its API
can host, a synthesis plan assigns every API a number of instances
to generate at its level, so that the APIs able to host L4--L5
receive the largest shares. For each planned instance, the
generation pipeline produces an instruction, an upstream-tool
context, and the target parameters. The result is kept only if it
passes a structural check: the required fields are present, the gold
parameters validate against the JSON Schema, and the level label is
legal and consistent with the actual nesting depth.

\begin{table}[t]
\centering

\begin{minipage}[c]{\columnwidth}
\begin{minipage}[c]{0.60\linewidth}
\raggedright
\small
\begin{tabular}{@{}l@{\hspace{3pt}}r@{}}
\hline
\multicolumn{2}{@{}l@{}}{\texttt{grade(d, c, t, e):}} \\
\texttt{~if d=0 and c=0 and t=0:} & \texttt{L1} \\
\texttt{~elif d$\le$1 and c=0} & \\
\texttt{~~~and t$\le$1 and e=0:} & \texttt{L2} \\
\texttt{~elif d$\le$2 and c=0:} & \texttt{L3} \\
\texttt{~elif d$\le$3 and c$\ge$1:} & \texttt{L4} \\
\texttt{~else:} & \texttt{L5} \\
\hline
\end{tabular}
\end{minipage}%
\hfill
\begin{minipage}[c]{0.40\linewidth}
\raggedleft
\small
\setlength{\tabcolsep}{3pt}
\begin{tabular}{lrr}
\hline
Level & Inst. & Share \\
\hline
L1 & 102 & 10.0\% \\
L2 & 198 & 19.4\% \\
L3 & 208 & 20.4\% \\
L4 & 256 & 25.0\% \\
L5 & 258 & 25.2\% \\
\hline
\textbf{Total} & \textbf{1{,}022} & 100\% \\
\hline
\end{tabular}
\end{minipage}
\end{minipage}
\caption{The difficulty grading rule (left) and the resulting
instance distribution (right).}
\label{tab:bench-distribution}
\end{table}

\section{Experiments}
\label{sec:exp}

\begin{table*}[t]
\centering
\small
\setlength{\tabcolsep}{3pt}
\begin{tabular*}{\textwidth}{@{\extracolsep{\fill}}cc *{14}{>{\centering\arraybackslash}p{21pt}}}
\toprule
\multirow{2}{*}{Model} & \multirow{2}{*}{Method} & \multicolumn{2}{c}{\textsc{ParamBench}} & \multicolumn{2}{c}{NESTFUL} & \multicolumn{2}{c}{Seal-Tools} & \multicolumn{2}{c}{xLAM} & \multicolumn{2}{c}{BFCL} & \multicolumn{2}{c}{API-Bank} & \multicolumn{2}{c}{CFB} \\
\cmidrule(lr){3-4}\cmidrule(lr){5-6}\cmidrule(lr){7-8}\cmidrule(lr){9-10}\cmidrule(lr){11-12}\cmidrule(lr){13-14}\cmidrule(lr){15-16}
 & & EM & F1 & EM & F1 & EM & F1 & EM & F1 & EM & F1 & EM & F1 & EM & F1 \\
\midrule
\multirow{6}{*}{Qwen3-8B}
 & 0-shot       & 14.7 & 63.7 & 11.4 & 20.5 & 58.4 & 72.8 & 56.2 & 64.3 & 46.5 & 70.1 & 44.7 & 57.7 & 23.9 & 39.7 \\
 & SeedSFT      & 23.0 & 70.3 & 35.8 & 47.9 & 79.2 & 89.6 & 81.8 & 89.6 & 69.2 & 84.6 & 77.0 & 88.1 & 41.6 & 53.2 \\
 & SelfTrain    & 17.8 & 69.9 & 34.6 & 47.2 & 81.0 & 89.8 & 81.4 & 89.1 & 71.6 & 88.1 & 77.9 & 88.1 & 43.2 & 53.9 \\
 & LogprobTrain & 23.9 & 70.2 & 36.4 & 48.8 & 52.4 & 57.6 & 73.4 & 78.8 & 49.5 & 61.3 & 54.9 & 63.9 & 38.9 & 50.8 \\
 & ConsistTrain & 13.0 & 36.5 & 33.4 & 43.9 & 42.4 & 47.2 & 72.8 & 79.0 & 59.2 & 72.9 & 69.0 & 77.9 & 44.2 & 53.9 \\
 & PBT          & \textbf{28.7} & \textbf{76.4} & \textbf{38.4} & \textbf{51.0} & \textbf{82.0} & \textbf{91.0} & \textbf{82.2} & \textbf{89.9} & \textbf{73.1} & \textbf{88.9} & \textbf{80.5} & \textbf{89.2} & \textbf{44.9} & \textbf{55.8} \\
\midrule
\multirow{6}{*}{Qwen3-14B}
 & 0-shot       & 24.9 & 69.2 & 4.0 & 6.7 & 30.0 & 35.5 & 16.8 & 20.1 & 24.9 & 39.2 & 44.7 & 54.9 & 15.6 & 29.3 \\
 & SeedSFT      & 26.6 & 77.2 & 41.8 & 53.1 & 69.0 & 81.5 & 71.5 & 81.2 & 70.8 & 86.2 & 77.0 & 87.0 & 36.3 & 50.0 \\
 & SelfTrain    & 35.2 & 79.0 & 41.4 & 51.7 & 79.9 & 89.6 & 79.2 & 87.5 & 72.7 & 87.6 & 76.1 & 87.2 & 41.0 & 53.0 \\
 & LogprobTrain & 8.9 & 27.5 & 15.6 & 19.4 & 79.0 & 88.2 & 77.2 & 86.0 & 71.1 & 86.9 & 7.1 & 7.8 & 12.6 & 14.2 \\
 & ConsistTrain & 19.5 & 56.8 & 24.6 & 30.1 & 80.0 & 89.7 & 79.2 & 87.2 & 52.6 & 63.8 & 77.4 & 87.4 & 12.1 & 17.6 \\
 & PBT          & \textbf{35.8} & \textbf{80.4} & \textbf{42.6} & \textbf{53.3} & \textbf{80.6} & \textbf{89.8} & \textbf{81.0} & \textbf{88.8} & \textbf{77.1} & \textbf{90.2} & \textbf{80.1} & \textbf{88.3} & \textbf{44.0} & \textbf{54.3} \\
\midrule
\multirow{6}{*}{Gemma-4-12B}
 & 0-shot       & 6.8 & 30.2 & 4.6 & 8.5 & 13.2 & 21.7 & 0.4 & 1.0 & 5.7 & 7.2 & 31.9 & 43.5 & 11.6 & 27.2 \\
 & SeedSFT      & 19.5 & 62.6 & 21.4 & 29.7 & 68.6 & 77.0 & 35.6 & 42.6 & 63.1 & 84.2 & 71.7 & 84.2 & 22.7 & 30.5 \\
 & SelfTrain    & 19.8 & 58.1 & 13.6 & 16.0 & 76.0 & 84.4 & 51.2 & 59.6 & 62.1 & 76.9 & 72.1 & 82.7 & 4.9 & 6.4 \\
 & LogprobTrain & 18.4 & 54.4 & 14.0 & 18.8 & 72.0 & 80.6 & 73.4 & 81.7 & 71.6 & 86.5 & 71.2 & 82.5 & 4.5 & 5.8 \\
 & ConsistTrain & 18.1 & 60.7 & \textbf{24.6} & 34.2 & 77.0 & 85.6 & 47.0 & 55.9 & 71.7 & 86.5 & 76.1 & 85.3 & 4.9 & 7.6 \\
 & PBT          & \textbf{23.9} & \textbf{73.8} & 23.8 & \textbf{35.2} & \textbf{78.8} & \textbf{88.7} & \textbf{74.2} & \textbf{82.3} & \textbf{73.1} & \textbf{87.6} & \textbf{79.2} & \textbf{87.5} & \textbf{30.6} & \textbf{40.2} \\
\midrule
\multirow{6}{*}{Ministral-3-8B}
 & 0-shot       & 14.0 & 55.3 & 10.8 & 17.7 & 38.0 & 51.3 & 6.0 & 6.7 & 8.2 & 11.1 & 67.7 & 79.5 & 1.9 & 3.1 \\
 & SeedSFT      & 25.9 & 72.5 & 17.4 & 19.8 & 66.0 & 75.6 & 71.8 & 77.9 & 67.5 & 85.2 & 78.8 & 87.5 & 32.0 & 46.6 \\
 & SelfTrain    & 27.0 & 75.8 & 4.0 & 4.0 & 78.0 & 87.4 & 72.8 & 80.2 & 71.8 & 86.2 & 83.6 & 90.3 & 41.4 & 53.8 \\
 & LogprobTrain & 29.0 & 76.3 & 3.8 & 4.5 & 77.0 & 85.8 & 42.1 & 45.5 & 59.5 & 71.5 & 76.5 & 83.5 & 40.1 & 51.5 \\
 & ConsistTrain & 26.3 & 75.4 & 0.6 & 1.6 & 75.0 & 83.8 & 69.8 & 76.0 & 70.1 & 85.7 & 84.1 & 91.5 & 39.9 & 53.2 \\
 & PBT          & \textbf{32.1} & \textbf{79.3} & \textbf{21.8} & \textbf{26.9} & \textbf{80.4} & \textbf{90.5} & \textbf{80.4} & \textbf{88.0} & \textbf{74.9} & \textbf{90.4} & \textbf{85.0} & \textbf{92.2} & \textbf{44.0} & \textbf{54.4} \\
\midrule
\multirow{6}{*}{Llama-3.1-8B}
 & 0-shot       & 0.0 & 0.8 & 1.6 & 2.5 & 1.2 & 1.3 & 20.4 & 22.5 & 11.0 & 16.0 & 12.4 & 13.3 & 4.4 & 4.4 \\
 & SeedSFT      & 10.9 & 46.3 & 14.6 & 18.0 & 66.4 & 79.0 & 72.8 & 80.6 & 67.7 & 84.8 & 75.7 & 87.6 & 36.5 & 50.3 \\
 & SelfTrain    & 22.9 & 71.4 & 23.8 & 33.5 & 73.2 & 82.3 & 72.0 & 78.9 & 64.8 & 80.5 & 80.5 & 88.9 & 37.9 & 52.0 \\
 & LogprobTrain & 23.9 & 74.2 & 25.4 & 36.0 & 70.6 & 80.2 & 1.9 & 3.1 & 66.5 & 82.2 & 68.1 & 74.9 & 37.3 & 52.4 \\
 & ConsistTrain & 27.6 & 75.0 & 22.4 & 31.6 & 57.6 & 64.6 & 47.2 & 52.6 & 70.1 & 85.7 & 82.3 & 90.0 & 38.3 & 52.4 \\
 & PBT          & \textbf{31.0} & \textbf{76.2} & \textbf{29.8} & \textbf{43.1} & \textbf{73.8} & \textbf{86.7} & \textbf{78.6} & \textbf{85.5} & \textbf{72.0} & \textbf{86.8} & \textbf{84.5} & \textbf{90.7} & \textbf{43.2} & \textbf{53.8} \\
\bottomrule
\end{tabular*}
\caption{Exact match (EM) and field-level F1 of the 6 methods on
\textsc{ParamBench} (PB) and 6 external tool-use
benchmarks (CFB = ComplexFuncBench) with all models in
their base variants. Bold marks the best value in each column of a
model block.}
\label{tab:main}
\end{table*}

In this section, we evaluate PBT and PGR from multiple perspectives.
The study is organized around four research questions (RQs).
\begin{itemize}
\item[\textbf{RQ1}] Does generating SFT data with the probe signal,
as in PBT, lead to better parameter generation?
\item[\textbf{RQ2}] How much do PBT and PGR improve parameter
generation over other leading tool-use methods?
\item[\textbf{RQ3}] How should the PGR strategy be chosen, and how
much can it improve performance at inference time?
\item[\textbf{RQ4}] How do PBT and PGR perform across the five
difficulty levels, L1 to L5?
\end{itemize}

\subsection{Experiment Settings}
\label{subsec:settings}

\paragraph{Datasets.} \textsc{ParamBench} is the in-domain benchmark.
We add 6 external tool-use benchmarks: NESTFUL~\citep{nestful-emnlp25},
Seal-Tools~\citep{sealtools-nlpcc24}, xLAM~\citep{apigen-neurips24},
BFCL~\citep{bfcl-icml25}, API-Bank~\citep{apibank-emnlp23}, and
ComplexFuncBench~\citep{complexfuncbench-arxiv25}.
\textsc{ParamBench} is split by API: 57 APIs (729 instances) form the
training set, and 24 unseen APIs (293 instances) form the test set.
Each external benchmark is split into a training set and a test set,
using the official split where available.

\paragraph{Adaptation.} The external benchmarks were built to test
which tool to pick and how to chain calls. We convert every instance
into a per-call record: the input gives the instruction, the spec of
the target tool, and the outputs of the earlier calls in the
chain; the model must output only the parameter object of the current
call. Tool selection is taken out of the task, so what remains is
parameter generation. When a benchmark accepts several values for a
field, matching any of them counts (BFCL). After adaptation, each
dataset is treated as its own domain: the probe and the
hyperparameters come from its own training set only.

\begin{table*}[t]
\centering
\small
\setlength{\tabcolsep}{3pt}
\begin{tabular*}{\textwidth}{@{\extracolsep{\fill}}l *{14}{>{\centering\arraybackslash}p{21pt}}}
\toprule
\multirow{2}{*}{Model} & \multicolumn{2}{c}{\textsc{ParamBench}} & \multicolumn{2}{c}{NESTFUL} & \multicolumn{2}{c}{Seal-Tools} & \multicolumn{2}{c}{xLAM} & \multicolumn{2}{c}{BFCL} & \multicolumn{2}{c}{API-Bank} & \multicolumn{2}{c}{CFB} \\
\cmidrule(lr){2-3}\cmidrule(lr){4-5}\cmidrule(lr){6-7}\cmidrule(lr){8-9}\cmidrule(lr){10-11}\cmidrule(lr){12-13}\cmidrule(lr){14-15}
 & EM & F1 & EM & F1 & EM & F1 & EM & F1 & EM & F1 & EM & F1 & EM & F1 \\
\midrule
 Llama-xLAM-2-8b-fc-r & 25.6 & 73.8 & 7.2 & 9.2 & 52.6 & 79.0 & 74.2 & 82.5 & 40.4 & 73.5 & 21.9 & 30.4 & 34.7 & 48.0 \\
 Hammer2.1-7b & \textbf{33.1} & 78.3 & 7.8 & 9.3 & 60.6 & 81.7 & 74.2 & 81.1 & 38.8 & 72.4 & 37.8 & 54.5 & 33.1 & 47.3 \\
 ToolACE-2.5-Llama-3.1-8B & 25.6 & 75.3 & 6.8 & 7.8 & 53.0 & 79.3 & 68.4 & 77.3 & 43.1 & 77.5 & 42.4 & 60.6 & 37.8 & 50.7 \\
 watt-tool-8B & 26.6 & 61.0 & 6.8 & 8.7 & 51.0 & 76.6 & 68.8 & 78.0 & 42.4 & 75.5 & 33.7 & 48.1 & 32.6 & 44.9 \\
 \textbf{Qwen3-8B + PBT+PGR (ours)} & \textbf{33.1} & \textbf{79.8} & \textbf{38.8} & \textbf{50.9} & \textbf{82.4} & \textbf{91.2} & \textbf{83.2} & \textbf{90.6} & \textbf{73.6} & \textbf{89.1} & \textbf{81.0} & \textbf{89.1} & \textbf{45.7} & \textbf{56.1} \\
\bottomrule
\end{tabular*}
\caption{Parameter generation against 4 open tool-use models,
under the per-call protocol;
the open models use 3-shot prompting and their native
function-calling format. Bold marks the best value in each column.}
\label{tab:sota-frontier}
\end{table*}

\begin{figure*}[t]
\centering
\begin{minipage}[b]{0.32\textwidth}
\centering
\includegraphics[width=\linewidth]{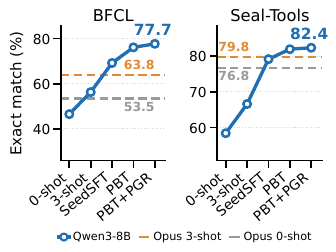}
\caption{Exact match of Qwen3-8B climbing from 0-shot to PBT+PGR, against Claude
Opus 4.7.}
\label{fig:ladder}
\end{minipage}\hfill
\begin{minipage}[b]{0.32\textwidth}
\centering
\includegraphics[width=\linewidth]{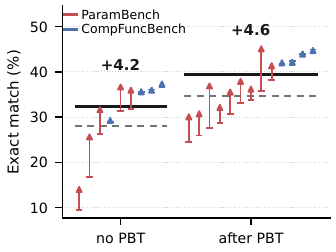}
\caption{PGR with and without PBT; dashed lines give the group mean
before reranking, solid lines after.}
\label{fig:pgrgain}
\end{minipage}\hfill
\begin{minipage}[b]{0.32\textwidth}
\centering
\includegraphics[width=\linewidth]{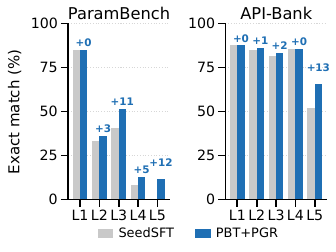}
\caption{Exact match of Qwen3-8B by difficulty level. Labels give the
gain of PBT+PGR over SeedSFT at each level.}
\label{fig:levels}
\end{minipage}
\end{figure*}

\paragraph{Models.} We use 5 open models from 4 families: Qwen3 (8B
and 14B), Gemma-4-12B, Ministral-3-8B, and Llama-3.1-8B; the main
analysis is on their base variants. For comparison, we also measure
4 frontier models, Claude Opus 4.7, GPT-5.4, DeepSeek-V4-Pro, and
Qwen-3.6-Plus, under 0-shot and 3-shot prompting.

\paragraph{Metrics.} We use two metrics. Exact match (EM) counts a
call as correct only if the generated parameter object equals the
gold one exactly, so a single wrong field makes the whole call
wrong. Field-level F1 gives partial credit for the fields that are
right: the predicted and the gold parameter objects are flattened
into $<$path, value$>$ pairs with every field weighted equally, and
F1 is the harmonic mean of precision and recall over the two sets.

\subsection{Results}
\label{subsec:results}

\paragraph{RQ1: Probe-filtered SFT.}
Table~\ref{tab:main} compares PBT with 5 baselines over 5 base models
and 7 datasets. 0-shot prompts the base model, and SeedSFT
fine-tunes it on a seed set of gold-labeled instances. Each remaining
method retrains the base model on the seed set plus self-generated
answers, and they differ only in the selection signal: SelfTrain
keeps every answer~\citep{scudder-tit65,selftrain-iclr20};
LogprobTrain and ConsistTrain keep the same number of answers as PBT,
selected by mean log-probability or by majority
vote~\citep{selfconsistency-iclr23}; and PBT selects by the probe
signal of Section~\ref{sec:method}.

0-shot shows why fine-tuning is necessary: a base model often
cannot even emit a well-formed parameter object, and it averages only
19.7 EM and 30.5 F1. SeedSFT fixes most of the format errors and
lifts the averages to 51.6 EM and 67.5 F1. The three self-training
baselines select data by unreliable signals, so the added data is noisy
and the outcome is unstable: on average LogprobTrain reaches only
44.2 EM, ConsistTrain 48.9, and SelfTrain 53.9. PBT is the only
method that improves steadily: it is above SeedSFT in all 35
model--dataset pairs, and it raises the averages to 59.6 EM
and 75.1 F1, a gain of 8.0 EM points over SeedSFT.

\paragraph{RQ2: PBT and PGR versus other tool-use methods.}
Table~\ref{tab:sota-frontier} compares our model, Qwen3-8B with PBT
and PGR, against 4 open tool-use models at the same scale:
Llama-xLAM-2-8b-fc-r~\citep{apigenmt-neurips25},
Hammer2.1-7b~\citep{hammer-iclr25},
ToolACE-2.5-Llama-3.1-8B~\citep{toolace-iclr25}, and
watt-tool-8B~\citep{watttool-hf24}. Our model is the best in every
column, and it averages 62.5 EM over the 7 datasets, 21.8 points
above the strongest of the other 4 models (Hammer2.1-7b, 40.8).

We further compare our model with the 4 frontier models, Claude
Opus 4.7, GPT-5.4, DeepSeek-V4-Pro, and Qwen-3.6-Plus, under 3-shot
prompting: our model reaches the frontier level on all 7 datasets
and surpasses all 4 frontier models on 3 datasets.
Figure~\ref{fig:ladder} shows this climb on BFCL and Seal-Tools:
Qwen3-8B starts far below Claude Opus 4.7, gains at every stage from
prompting to SeedSFT and PBT, and ends above Opus 4.7 on both
datasets. The 0-shot setting and the full frontier panel are in the
appendix.

\paragraph{RQ3: PGR strategy choice and inference-time gain.}
PGR samples a pool of candidate calls for each instruction and
picks the final call from this pool. Section~\ref{subsec:rerank}
gives 3 families of decision strategies, and each fits a different
failure mode. \emph{Candidate-level} strategies fit when the pool
usually holds one fully correct call; \emph{field-level} strategies
fit when no candidate is right as a whole but every field is right
in some candidate; \emph{field-set} strategies fit when the greedy
values are right but the call adds fields it should not. A base
model gets many values wrong, so there is much to rebuild: the
field-level strategy, which takes each field from its best
candidate, gains 7.5 EM over greedy on the base Qwen3-14B. After
PBT the greedy call is already right most of the time, so the
conservative strategy that stays close to it works best, lifting
\textsc{ParamBench} EM from 35.8 to 44.4. In addition, as shown in Figure~\ref{fig:pgrgain},
PGR does not depend on PBT and works on its own: applied directly
to models that were never fine-tuned, as in the left half of the
figure, it lifts exact match by 4.2 points on average, and applied
after PBT, as in the right half, it adds 4.6 points.

\paragraph{RQ4: Performance across difficulty levels.}
Figure~\ref{fig:levels} splits the exact match of Qwen3-8B on
\textsc{ParamBench} and API-Bank by the five difficulty levels,
comparing SeedSFT with PBT+PGR. At the easy end the two methods are
close: L1 is already solved well by SeedSFT alone, and L2 gains a
little. The gains concentrate at the hard levels. On
\textsc{ParamBench}, PBT+PGR adds 11 points at L3, 5 points at L4,
and 12 points at L5, where SeedSFT is close to zero; on API-Bank,
the first four levels already sit above 80 EM, and the gain appears
at L5, which rises by 13 points. The hard levels are exactly where
deep nesting, conditional dependencies, and cross-call derivation
appear, so the improvement there shows that PBT and PGR learn part
of what makes parameter generation difficult, instead of only
polishing what the seed model already handles.

\section{Conclusion}
\label{sec:conclusion}

This paper puts tool-call parameter generation at the center of
tool-use research, tracing its difficulty to deep nesting,
conditional dependency, and cross-call derivation. We show that a
linear probe on the hidden state, read before a parameter object is written,
predicts whether the value will be correct; PBT uses the signal to
filter self-generated training data, and PGR to rerank sampled
candidates. For evaluation, we release \textsc{ParamBench}, a
benchmark built from real cloud-network agent traces and graded
into five difficulty levels. Extensive experiments across open
models and external benchmarks validate the effectiveness of both
PBT and PGR.

\paragraph{Limitations.}
First, \textsc{ParamBench} covers only one scenario: its APIs and
traces all come from cloud networking, although the difficulty
scale carries over to other domains. Second, the probe does not
transfer: tied to one dataset, model, and sampling temperature, it
degrades outside that setting, so a new probe must be fit for every new domain. Third, PBT and PGR depend on a small labeled set and a
pool of unlabeled instructions; more general methods for
strengthening parameter generation remain to be explored.

\bibliography{refs}

\clearpage

\appendix
\renewcommand{\topfraction}{0.95}
\renewcommand{\floatpagefraction}{0.9}
\renewcommand{\dbltopfraction}{0.99}
\renewcommand{\textfraction}{0.05}
\setcounter{topnumber}{3}
\setcounter{dbltopnumber}{2}
\renewcommand{\dblfloatpagefraction}{0.95}
\makeatletter
\setlength{\@fptop}{0pt}
\setlength{\@fpsep}{14pt}
\setlength{\@fpbot}{0pt plus 1fil}
\setlength{\@dblfptop}{0pt}
\setlength{\@dblfpsep}{12pt plus 1fil}
\setlength{\@dblfpbot}{0pt}
\makeatother
\linespread{1.115}\selectfont
\renewcommand{\arraystretch}{1.27}
\setlength{\abovecaptionskip}{8pt}
\setlength{\belowcaptionskip}{2pt}

\section{ParamBench Details}
\label{app:parambench}

\subsection{Positioning Against Existing Benchmarks}
\label{app:positioning}

Table~\ref{tab:bench-positioning} contrasts \textsc{ParamBench} with the
representative tool-use benchmarks discussed in
Section~\ref{sec:related}.

\begin{table*}[t]
\centering
\small
\setlength{\tabcolsep}{6pt}
\begin{tabular}{lcccc}
\hline
Dimension & ToolBench & API-Bank & NESTFUL / Seal-Tools & \textsc{ParamBench} \\
\hline
Focus           & multi-tool & API use   & nested calls & \textbf{param. values} \\
Param.\ nesting & $\le 1$    & $\le 1$   & flat (call-level chains) & \textbf{0--5} \\
Cond.\ deps     & --         & --        & --           & \checkmark \\
Cross-call      & implicit   & --        & chained      & \checkmark explicit \\
Difficulty grade& --         & --        & binary tag (Seal) & L1--L5 \\
\hline
\end{tabular}
\caption{Positioning of \textsc{ParamBench} against representative
tool-use benchmarks: ToolBench~\citep{toolllm-iclr24},
API-Bank~\citep{apibank-emnlp23}, NESTFUL~\citep{nestful-emnlp25}, and
Seal-Tools~\citep{sealtools-nlpcc24}. ``Param.\ nesting'' = nesting
depth of a single call's parameter object; NESTFUL and Seal-Tools nest
at the call level (one call's output feeds a later call) while their
per-call parameters stay largely flat. ``Cond.\ deps'' = inter-field
conditional dependencies; ``Cross-call'' = a downstream parameter
explicitly derived from an upstream output. NESTFUL and Seal-Tools also serve as the
external evaluation domains in Section~\ref{sec:exp}.}
\label{tab:bench-positioning}
\end{table*}

\subsection{Construction and Validation}
\label{app:construction}

Trace-extracted instances are produced by the converter described in
Section~\ref{sec:parambench}. The converter walks each verified
diagnosis trace and turns every executable step into one instance:
the upstream context keeps only the fields that the call actually
reads, and the step's verified parameters become the gold answer.
Before release, every instance passed two checks. First, the gold
answer must validate against the frozen schema of its tool. Second,
all organization-specific values are replaced with anonymized
aliases under one consistent mapping covering resource ids (prefix
and length kept), region ids and display names, account ids,
domains, IP addresses, instance specs, metric namespaces and keys,
runbook and scenario codes, and uniformly shifted timestamps. A
value maps to the same alias in the instruction, the context, and
the gold answer, preserving cross-call transfers and exact-match
scoring.

Synthesized instances follow a per-schema plan. Because the
complexity profile of a schema determines which difficulty levels its
API can host, the plan assigns the largest shares to the APIs that
can host L4 and L5. A synthesized instance is kept only when three
conditions hold: its required fields are present, its gold parameters
validate against the schema, and its level label matches the
structure that was actually generated.

Each instance names its target tool, and every name resolves
against the frozen pool of 81 cloud-network API schemas.

\subsection{A Complete Example Instance}
\label{app:example}

The listing below shows an L5 instance in full. The
target call is \texttt{DescribeDBInstancePerformance}. Three parts of
the gold answer cannot be copied from the instruction. The
\texttt{DBInstanceId} must be taken from the output of an earlier
\texttt{DescribeDBInstances} call. The \texttt{Key} field must
compose 3 metric names from the reported symptoms. The time window
must be converted from Beijing time in the instruction to UTC, and
its start is anchored by an event detail returned by an earlier
\texttt{DescribeEvents} call.

\begin{figure}[t]
\centering
\begin{minipage}{0.98\columnwidth}
\small
\begin{verbatim}
{"sample_id": "PB-eval2-gen-L5-0004",
 "tool": "DescribeDBInstancePerformance",
 "instruction": "[translated from Chinese]
   An RDS MySQL 8.0 instance reports
   intermittent slow-query alerts. Pull the
   QPS/TPS, slow-log count, and memory/CPU
   curves for 03:00-09:00 Beijing time
   today.",
 "golden_params": {
   "DBInstanceId": "rm-aaaaaaaaaaaaaaaxe",
   "Key": "DB_QPSTPS,DB_SlowLogs,
           DB_MemCpuUsage",
   "StartTime": "2026-06-03T19:00Z",
   "EndTime": "2026-06-04T01:00Z"},
 "complexity_labels": {"level": "L5",
   "num_upstream_transfers": 2},
 "data_flow_context": {"upstream": [
   {"tool": "DescribeDBInstances",
    "output": {"DBInstanceId":
      "rm-aaaaaaaaaaaaaaaxe",
      "Engine": "MySQL", ...}},
   {"tool": "DescribeEvents",
    "output": {"EventName":
      "SlowQueryThresholdExceeded",
      "EventTime": "2026-06-04T00:17Z",
      "Detail": "[translated] window
        started at 2026-06-03T19:02Z",
      ...}}]}}
\end{verbatim}
\end{minipage}
\caption*{An L5 \textsc{ParamBench} instance, abridged. The
released record renders these blocks inside chat messages; the
listing regroups them, names the API behind each upstream output,
translates Chinese free text, and truncates long outputs.
Identifier and value fields match the released record.}
\end{figure}

\section{Experimental Setup Details}
\label{app:setup}

\subsection{External Benchmark Adaptation}
\label{app:adaptation}

Every external instance is converted into the per-call record
described in Section~\ref{sec:exp}. A record contains the user
query, the name and JSON spec of the one tool to call, the outputs
of previously executed steps (each truncated to 280 characters), and
the gold arguments. The conversion rules differ from source to
source. BFCL parallel-call categories are skipped, because a
parallel call has no single current tool. xLAM calls whose tool list
contains duplicate names are dropped. NESTFUL is split at the
sequence level with an 80/20 ratio, and its test records are drawn
only from the held-out sequences. Seal-Tools uses its official
split. For xLAM, the training pool excludes every sample whose
source id appears in the test file, and a split audit confirms that
the train and test source ids do not overlap.

Table~\ref{tab:test-universes} lists the frozen test sets. For the 3
benchmarks used only for evaluation (BFCL, API-Bank,
ComplexFuncBench), the few-shot examples (30 per dataset, fixed seed)
are carved out of the pool first, the 150 labeled and 250 unlabeled
training records are drawn next, and the remainder is the frozen
evaluation set.

The frontier panel of Table~\ref{tab:frontier-full} was run on a
different record subset for these 3 benchmarks (500/456/500 records
against our 770/226/770). Both panels are therefore rescored on the
intersection of the two subsets ($n = 318/226/318$) from per-record
outputs. On API-Bank the intersection equals our full set, and the
rescored PBT row reproduces Table~\ref{tab:main} exactly.

\begin{table}[t]
\centering
\small
\setlength{\tabcolsep}{5pt}
\begin{tabular}{@{}lr@{}}
\toprule
Dataset & Test records \\
\midrule
\textsc{ParamBench} & 293 \\
NESTFUL & 500 \\
Seal-Tools & 500 \\
xLAM & 500 (drawn from a 1{,}200-record pool) \\
BFCL & 770 \\
API-Bank & 226 \\
ComplexFuncBench & 770 \\
\bottomrule
\end{tabular}
\caption{Frozen test sets after per-call conversion.}
\label{tab:test-universes}
\end{table}

\subsection{Prompt Templates}
\label{app:prompts}

Two templates are used. The \textsc{ParamBench} runner uses the
system prompt of Figure~\ref{fig:prompt-pb} (English translation;
the released records embed the Chinese original). The user message
gives the instruction, the target API schema, and the upstream
context as titled blocks. The prompt requests a fixed four-key
reasoning object whose final \texttt{params} key carries the
parameter object scored against the gold. In the 3-shot setting,
the examples come from difficulty levels other than the target's.

\begin{figure}[t]
\centering
\includegraphics[width=0.95\columnwidth]{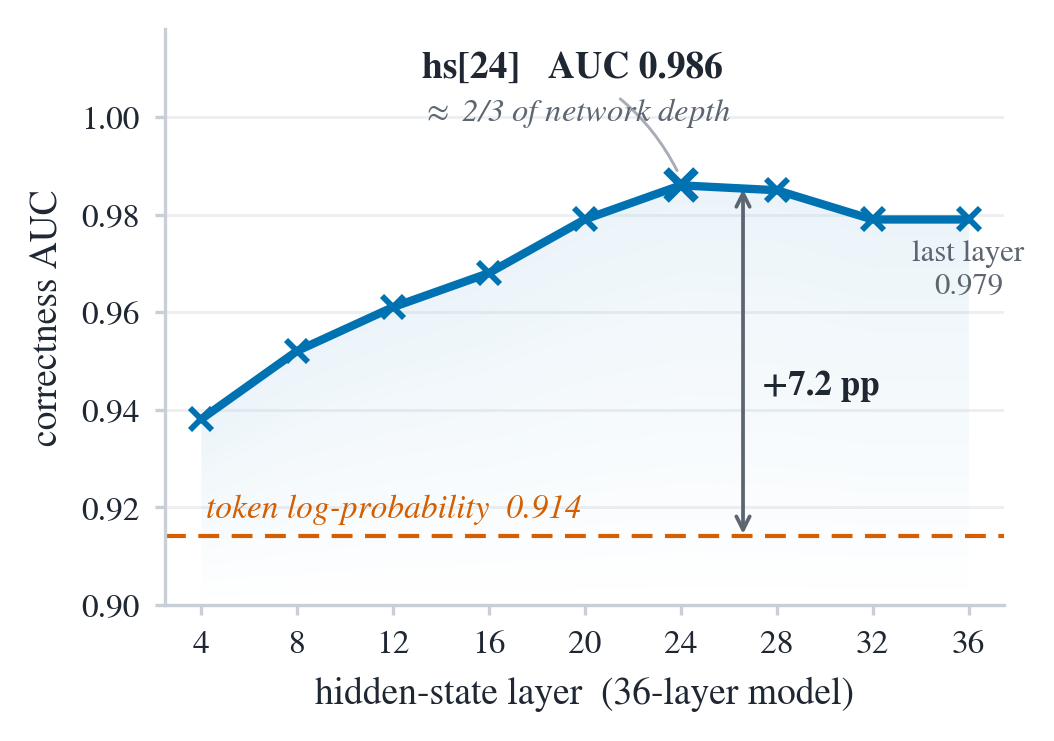}
\caption{Per-layer AUC of the correctness probe. The best layer sits
at about two-thirds of network depth.}
\label{fig:auc}
\end{figure}

\begin{figure}[t]
\centering
\begin{minipage}{0.98\columnwidth}
\small
\begin{verbatim}
You are a Platform-X OpenAPI expert. Given
a natural-language instruction, the target
API schema, and the upstream tool-output
context, reason in three steps (schema
structure analysis -> parameter source
judgment -> layer-by-layer construction of
the nested parameters) and produce the
parameter JSON for a direct call.

Output strictly the following structure
(all four top-level keys required):
{
  "step1_schema_analysis": "...",
  "step2_source_judgment": [
    {"param_path": "...",
     "source_kind": "literal|from_context
                     |from_step|derive",
     "source_detail": "..."}
  ],
  "step3_param_construction": "...",
  "params": { ... final JSON ... }
}

Do not return markdown fences. Do not add
explanatory text.
\end{verbatim}
\end{minipage}
\caption{System prompt of the \textsc{ParamBench} runner, in
English translation. The released records embed the Chinese
original.}
\label{fig:prompt-pb}
\par\vspace{6pt}
\begin{minipage}{0.98\columnwidth}
\small
\begin{verbatim}
You are an API function-calling expert.
Given a user query, the JSON spec of ONE
tool to call, and the outputs of previously
executed steps, produce ONLY the arguments
for the current tool call as a JSON object.
When an argument value must come from a
previous step's output, use the same
reference convention shown in the
examples/context (e.g. a step label or the
literal value). Output strictly a JSON
object, no prose.
\end{verbatim}
\end{minipage}
\caption{System prompt of the external per-call template.}
\label{fig:prompt-ext}
\end{figure}

The 6 external benchmarks use the per-call template of
Figure~\ref{fig:prompt-ext}; the user message gives the user query, the spec of the tool to call now, and
the outputs of previously executed steps.

\subsection{Hyperparameters}
\label{app:hparams}

Table~\ref{tab:hyperparams} lists the final settings behind the
main tables. The learning rate and the number of epochs were chosen
per model family from $\{5\times10^{-4}, 1\times10^{-4}\}\times\{2,
4\}$ on a held-out 20\% split of the seed set: Gemma-4 uses
$1\times10^{-4}$; Llama-3.1 uses $1\times10^{-4}$ and Ministral-3
uses 2 epochs on NESTFUL; all other cells use $5\times10^{-4}$ and
4 epochs. The PBT threshold $\tau$ was swept over $\{0.8, 0.9,
0.95\}$ on the training side. All remaining values were fixed in
advance and not searched.

\begin{table}[t]
\centering
\small
\setlength{\tabcolsep}{4pt}
\begin{tabular}{@{}ll@{}}
\toprule
Setting & Value \\
\midrule
\multicolumn{2}{@{}l}{\emph{LoRA fine-tuning (seed SFT and PBT retraining)}} \\
rank / alpha / dropout & 32 / 64 / 0.05 \\
target modules & q, k, v, o projections \\
learning rate & $5\times10^{-4}$, cosine, 3\% warmup \\
epochs & 4 \\
batch size & 1, gradient accumulation 8 \\
precision & bf16 \\
max sequence length & 2048 \\
\midrule
\multicolumn{2}{@{}l}{\emph{PBT}} \\
labeled + unlabeled & PB 364 + 365 \\
 & NF/Seal/xLAM 500 + 2000 \\
 & BFCL/API-Bank/CFB 150 + 250 \\
candidates per instruction & 1 greedy + 6 sampled \\
sampling & $T = 0.8$, top-$p$ 0.95 \\
threshold $\tau$ & 0.9 (NESTFUL 0.95) \\
rounds & 1 \\
\midrule
\multicolumn{2}{@{}l}{\emph{Probe}} \\
classifier & logistic regression, $C = 1$ \\
features & pre-value hidden state, standardized \\
layer & $\{4, 8, \ldots, 32, \mathrm{last}\}$, best CV AUC \\
training data & sampled candidates of the training set \\
\midrule
\multicolumn{2}{@{}l}{\emph{PGR}} \\
test-side pool & 1 greedy + 12 sampled, $T = 0.8$ \\
 & (models without fine-tuning: 1 + 8) \\
candidate score & mean per-parameter probe probability \\
strategy choice & 5-fold selection on the training set \\
\midrule
\multicolumn{2}{@{}l}{\emph{Evaluation}} \\
decoding & greedy, at most 640 new tokens \\
frontier and open panels & $T = 0$, 0-shot and 3-shot \\
\bottomrule
\end{tabular}
\caption{Final hyperparameters behind the main tables. Per-family
exceptions are listed in the text.}
\label{tab:hyperparams}
\end{table}

\subsection{Compute, Seeds, and Statistical Tests}
\label{app:repro}

\paragraph{Compute infrastructure.}
Training and evaluation ran on cloud GPU nodes, each with 8 NVIDIA
H20-3e GPUs (140 GB memory per GPU), two 48-core Intel Xeon Platinum
8575C CPUs, and 2 TiB RAM, running Linux with CUDA 12.8. The result
campaigns used up to 4 nodes (32 GPUs). All jobs ran in Docker
containers with PyTorch 2.8, Transformers 4.57.1, and PEFT 0.19.1;
candidate pools were generated with Hugging Face \texttt{generate}
or vLLM under the same sampling settings. Some of the Ministral and
Gemma arms were trained on a university server with 8 NVIDIA
A100-SXM4 GPUs (40 GB). The 4 frontier models were accessed through
their public inference APIs.

\paragraph{Runs and seeds.}
Unless stated otherwise, every number in the main tables comes from
a single run with training seed 42. The seed 43/47/53 rows of
Table~\ref{tab:pgr-pbt} report seed variants separately. Candidate
sampling does not fix a generator seed, so sampled pools vary across
runs; the repeated \textsc{ParamBench} and ComplexFuncBench runs behind
the cells of Table~\ref{tab:pgr-bare} show the size of this
variation.

\paragraph{Statistical significance.}
Two tests support the main comparisons. Over the 35 model--dataset
pairs of Table~\ref{tab:main}, PBT is above SeedSFT in every pair,
and a two-sided Wilcoxon signed-rank test on the paired EM values
gives $p = 5.8\times10^{-11}$. At the instance level, paired exact
McNemar tests on shared test instances confirm individual
contrasts: for example, the Qwen3-14B \textsc{ParamBench} gain from
SeedSFT (26.6) to PBT (35.8) over the 293 test instances has
$p = 1.4\times10^{-5}$. All fine-tuning runs use a single seed, so
all tests are paired at the instance level.

\section{Probe Analysis}
\label{app:probe}

\subsection{Per-Layer AUC}
\label{app:probe-layer}

Figure~\ref{fig:auc} gives the per-layer AUC of the correctness probe of
Section~\ref{sec:method}. One probe is trained per layer of the grid in
Table~\ref{tab:hyperparams} on the hidden state at the decision
point of each parameter, and the layer that performs best sits at
roughly two-thirds of the network depth.

\subsection{Robustness Across Scales, Checkpoints, and Datasets}
\label{app:probe-robust}

Table~\ref{tab:probe-robust} reports the decision-point probe AUC on
\textsc{ParamBench} across 4 Qwen3 scales, before and after
fine-tuning. All 8 cells fall between 0.944 and 0.993, consistent to
rounding with the 0.93--0.99 range stated in
Section~\ref{sec:method}.
The before and after columns use different prompting protocols and
some before-cells rest on a few hundred parameters each, so the
cells are indicative rather than strictly comparable.

The signal is also stable in three further directions. Refitting
with 10 random seeds gives AUC $0.983 \pm 0.003$. A probe trained
on one checkpoint keeps 0.982 on the next (0.986 in its own
setting), and a 3-checkpoint transfer matrix stays within
0.948--0.986. Nor is it Qwen-specific: a Llama-3.1-8B probe reaches
0.963 on \textsc{ParamBench}. What it does not survive is a change
of dataset: NESTFUL AUC drops to 0.770 (Qwen3-8B) and 0.711
(Llama-3.1-8B), which is why each domain trains its own probe, as
stated in Section~\ref{sec:method}.

\begin{table}[t]
\centering
\small
\setlength{\tabcolsep}{6pt}
\begin{tabular}{@{}lcc@{}}
\toprule
Model & Before FT & After FT \\
\midrule
Qwen3-8B & 0.973 & 0.968 \\
Qwen3-14B & 0.982 & 0.993 \\
Qwen3-32B & 0.965 & 0.988 \\
Qwen3-30B-A3B & 0.944 & 0.974 \\
\bottomrule
\end{tabular}
\caption{Decision-point probe AUC on \textsc{ParamBench} across
model scales, before and after fine-tuning.}
\label{tab:probe-robust}
\end{table}

\section{Full Results}
\label{app:results}

\subsection{Open Tool-Use Models}
\label{app:open}

Table~\ref{tab:open-full} reports the full open tool-use panel behind
Table~\ref{tab:sota-frontier}: the 4 open tool-use models of the main
text plus Qwen3-8B prompted in its native function-calling format, at
the same 7 to 8B scale, 0-shot and 3-shot, against our Qwen3-8B under
both PBT and PBT+PGR. The main text keeps the 3-shot rows of the 4
tuned models and the PBT+PGR row.

Three patterns stand out in the full panel. First, few-shot
prompting moves the open models unevenly: 3 shots lift API-Bank by
11 to 13 points for xLAM-2-8B-fc-r, Hammer2.1-7B, and watt-tool-8B,
but leave their \textsc{ParamBench} and NESTFUL columns almost
where they were. Second, NESTFUL is the weakest column for every
open model: none of the 5 exceeds 10.8 EM, against 38.8 for our
model, and the distance comes from the cross-call derivations that
the per-call record makes explicit. Third, the native
function-calling variant of Qwen3-8B is the strongest open entry on
BFCL, at 53.9 EM 0-shot, yet it still trails our PBT model by about
20 points on that dataset, so the gap is not a matter of output
format alone.

\begin{table*}[t]
\centering
\small
\setlength{\tabcolsep}{3pt}
\begin{tabular*}{\textwidth}{@{\extracolsep{\fill}}cc *{14}{>{\centering\arraybackslash}p{21pt}}}
\toprule
\multirow{2}{*}{Model} & \multirow{2}{*}{Setting} & \multicolumn{2}{c}{\textsc{ParamBench}} & \multicolumn{2}{c}{NESTFUL} & \multicolumn{2}{c}{Seal-Tools} & \multicolumn{2}{c}{xLAM} & \multicolumn{2}{c}{BFCL} & \multicolumn{2}{c}{API-Bank} & \multicolumn{2}{c}{CFB} \\
\cmidrule(lr){3-4}\cmidrule(lr){5-6}\cmidrule(lr){7-8}\cmidrule(lr){9-10}\cmidrule(lr){11-12}\cmidrule(lr){13-14}\cmidrule(lr){15-16}
 & & EM & F1 & EM & F1 & EM & F1 & EM & F1 & EM & F1 & EM & F1 & EM & F1 \\
\midrule
 \multirow{2}{*}{xLAM-2-8B-fc-r} & 0-shot & 23.9 & 73.7 & 5.4 & 6.2 & 50.2 & 76.4 & 73.6 & 81.8 & 41.0 & 74.5 & 9.3 & 17.2 & 31.7 & 45.2 \\
  & 3-shot & 25.6 & 73.8 & 7.2 & 9.2 & 52.6 & 79.0 & 74.2 & 82.5 & 40.4 & 73.5 & 21.9 & 30.4 & 34.7 & 48.0 \\
 \multirow{2}{*}{Hammer2.1-7B} & 0-shot & \textbf{34.8} & 78.1 & 5.2 & 6.0 & 58.6 & 80.7 & 74.0 & 81.3 & 39.0 & 73.4 & 25.2 & 37.4 & 27.1 & 42.0 \\
  & 3-shot & 33.1 & 78.3 & 7.8 & 9.3 & 60.6 & 81.7 & 74.2 & 81.1 & 38.8 & 72.4 & 37.8 & 54.5 & 33.1 & 47.3 \\
 \multirow{2}{*}{ToolACE-2.5-8B} & 0-shot & 23.6 & 71.2 & 5.0 & 6.0 & 53.0 & 78.9 & 68.0 & 77.2 & 42.2 & 76.0 & 43.4 & 61.0 & 36.0 & 47.7 \\
  & 3-shot & 25.6 & 75.3 & 6.8 & 7.8 & 53.0 & 79.3 & 68.4 & 77.3 & 43.1 & 77.5 & 42.4 & 60.6 & 37.8 & 50.7 \\
 \multirow{2}{*}{watt-tool-8B} & 0-shot & 27.3 & 61.9 & 5.2 & 5.7 & 51.8 & 77.1 & 70.4 & 79.5 & 42.5 & 75.7 & 22.1 & 35.6 & 32.5 & 44.9 \\
  & 3-shot & 26.6 & 61.0 & 6.8 & 8.7 & 51.0 & 76.6 & 68.8 & 78.0 & 42.4 & 75.5 & 33.7 & 48.1 & 32.6 & 44.9 \\
 \multirow{2}{*}{Qwen3-8B (native FC)} & 0-shot & 25.9 & 75.3 & 5.4 & 6.3 & 52.2 & 79.0 & 69.2 & 78.1 & 53.9 & 81.1 & 36.7 & 50.5 & 33.8 & 48.9 \\
  & 3-shot & 23.2 & 74.9 & 10.8 & 12.5 & 52.2 & 78.9 & 70.0 & 79.0 & 53.2 & 80.4 & 41.8 & 58.7 & 36.6 & 51.1 \\
 \textbf{Ours (PBT)} &   & 28.7 & 76.4 & 38.4 & \textbf{51.0} & 82.0 & 91.0 & 82.2 & 89.9 & 73.1 & 88.9 & 80.5 & \textbf{89.2} & 44.9 & 55.8 \\
 \textbf{Ours (PBT+PGR)} &   & 33.1 & \textbf{79.8} & \textbf{38.8} & 50.9 & \textbf{82.4} & \textbf{91.2} & \textbf{83.2} & \textbf{90.6} & \textbf{73.6} & \textbf{89.1} & \textbf{81.0} & 89.1 & \textbf{45.7} & \textbf{56.1} \\
\bottomrule
\end{tabular*}
\caption{Open tool-use models on \textsc{ParamBench} and the 6
external benchmarks under the identical per-call protocol of
Section~\ref{sec:exp}, 0-shot and 3-shot, against Qwen3-8B trained
with PBT and reranked with PGR. All rows share identical record sets
on all 7 datasets, so every column is row-comparable. Bold marks the
best value in each column across the 12 rows.}
\label{tab:open-full}
\end{table*}

\begin{table*}[t]
\centering
\small
\setlength{\tabcolsep}{3pt}
\begin{tabular*}{\textwidth}{@{\extracolsep{\fill}}cc *{14}{>{\centering\arraybackslash}p{21pt}}}
\toprule
\multirow{2}{*}{Model} & \multirow{2}{*}{Setting} & \multicolumn{2}{c}{\textsc{ParamBench}} & \multicolumn{2}{c}{NESTFUL} & \multicolumn{2}{c}{Seal-Tools} & \multicolumn{2}{c}{xLAM} & \multicolumn{2}{c}{BFCL} & \multicolumn{2}{c}{API-Bank} & \multicolumn{2}{c}{CFB} \\
\cmidrule(lr){3-4}\cmidrule(lr){5-6}\cmidrule(lr){7-8}\cmidrule(lr){9-10}\cmidrule(lr){11-12}\cmidrule(lr){13-14}\cmidrule(lr){15-16}
 & & EM & F1 & EM & F1 & EM & F1 & EM & F1 & EM & F1 & EM & F1 & EM & F1 \\
\midrule
 \multirow{2}{*}{Claude Opus 4.7} & 0-shot & 29.4 & 77.2 & 33.4 & 44.8 & 76.8 & 87.3 & 81.6 & 89.2 & 53.5 & 83.1 & 69.0 & 82.9 & 34.6 & 50.8 \\
  & 3-shot & 34.5 & 78.5 & \textbf{42.0} & \textbf{51.3} & 79.8 & 89.5 & 83.6 & 90.6 & 63.8 & 85.9 & 71.2 & 84.7 & 45.3 & 56.1 \\
 \multirow{2}{*}{GPT-5.4} & 0-shot & 32.4 & 76.2 & 28.2 & 39.5 & 76.4 & 87.7 & 78.2 & 86.4 & 5.7 & 10.6 & 67.3 & 81.3 & 34.6 & 50.8 \\
  & 3-shot & \textbf{41.3} & \textbf{80.9} & 32.0 & 42.6 & 79.6 & 89.1 & 83.6 & \textbf{91.0} & 61.6 & 83.9 & 71.2 & 84.7 & 42.5 & 53.8 \\
 \multirow{2}{*}{DeepSeek-V4-Pro} & 0-shot & 36.5 & 73.8 & 27.8 & 38.1 & 76.6 & 88.2 & 82.6 & 89.6 & 50.6 & 81.8 & 69.5 & 83.0 & 43.1 & 55.3 \\
  & 3-shot & 38.9 & 78.9 & 36.6 & 47.1 & 79.0 & 88.7 & 83.4 & 90.5 & 63.5 & 85.8 & 72.1 & 85.4 & 45.9 & 56.4 \\
 \multirow{2}{*}{Qwen-3.6-Plus} & 0-shot & 39.2 & 67.5 & 27.4 & 38.9 & 77.6 & 88.0 & 84.0 & 90.0 & 54.1 & 82.5 & 70.8 & 84.4 & 43.4 & 55.5 \\
  & 3-shot & 39.2 & 65.7 & 31.0 & 42.8 & 79.0 & 88.4 & \textbf{84.8} & 90.5 & 63.2 & 86.3 & 72.6 & 85.7 & \textbf{46.5} & \textbf{57.2} \\
 \textbf{Ours (PBT)} &  & 28.7 & 76.4 & 38.4 & 51.0 & 82.0 & 91.0 & 82.2 & 89.9 & \textbf{76.4} & \textbf{89.7} & 80.5 & \textbf{89.2} & 44.7 & 56.3 \\
 \textbf{Ours (PBT+PGR)} &  & 33.1 & 79.8 & 38.8 & 50.9 & \textbf{82.4} & \textbf{91.2} & 83.2 & 90.6 & 75.5 & 89.6 & \textbf{81.0} & 89.1 & 42.5 & 54.0 \\
\bottomrule
\end{tabular*}
\caption{Frontier models on \textsc{ParamBench} and the 6 external
benchmarks under the identical per-call protocol of
Section~\ref{sec:exp}, 0-shot and 3-shot, against Qwen3-8B trained
with PBT and reranked with PGR. GPT-5.4 fails to emit parseable
calls 0-shot on BFCL and recovers with 3 shots. The frontier panel
was run on a different record subset for BFCL, API-Bank and
ComplexFuncBench, so on those 3 datasets both panels are scored here
on the intersection of the two subsets ($n = 318/226/318$),
recomputed from per-record outputs. On BFCL the Ours rows use the
13-candidate pool of Table~\ref{tab:hyperparams}, where reranking
does not help (76.4 to 75.5 EM); Figure~\ref{fig:ladder} draws a
49-candidate BFCL pool, about 4 times that budget, where PGR gains
1.6 EM (76.1 to 77.7) on the same intersection. Every column is therefore
row-comparable and bold marks the best value in each column across
all 10 rows.}
\label{tab:frontier-full}
\end{table*}

\subsection{Frontier Models}
\label{app:frontier}

Table~\ref{tab:frontier-full} reports the full frontier panel referenced
in Section~\ref{sec:exp}: 4 frontier models measured on all 7
datasets under the same adapted per-call protocol as the local models,
0-shot and 3-shot, against our Qwen3-8B under both PBT and PBT+PGR.

For the frontier models, the value of 3 shots is concentrated where
the output format is the obstacle. GPT-5.4 fails to emit parseable
calls on BFCL 0-shot (5.7 EM) and recovers to 61.6 with examples,
and the other 3 models gain 9 to 13 points on BFCL from the same
treatment. On the remaining datasets the panel moves by a few
points at most and clusters tightly: at 3 shots the 4 models sit
within about 2 points of one another on Seal-Tools, xLAM, and
API-Bank. The margin of our model over the panel is largest exactly
there: +12.6 EM on BFCL over the best 3-shot frontier score and
+7.9 on API-Bank, both reached by PBT alone, before any reranking.

\subsection{Probe-Guided Reranking, Per-Arm Results}
\label{app:pgr-arms}

Tables~\ref{tab:pgr-bare} and~\ref{tab:pgr-pbt} give the per-arm
reranking measurements behind Figure~\ref{fig:pgrgain}, both
recomputed from the archived candidate pools under the same 5-fold
protocol. Table~\ref{tab:pgr-bare} covers instruction-tuned models
without fine-tuning; its 9 cells average $+4.2$ EM, the left half
of the figure. The base-variant comparison quoted in the main text
comes from an earlier archived pool: on the base Qwen3-14B
\textsc{ParamBench} pool (greedy 28.3), field-level splice adds
$+7.5$ EM and candidate-level probe argmax $+6.1$, while ranking by
log-probability loses points. Table~\ref{tab:pgr-pbt} covers the
same 2 datasets after PBT; the 13 arms drawn in the figure average
$+4.6$ EM, its right half. For the Qwen3-14B-Base
\textsc{ParamBench} cell both archived pools are listed: the main
text quotes the n12 pool (35.8 to 44.4), the figure draws the RS
pool.

The two sides of the figure behave differently. Without
fine-tuning, the gain tracks the weakness of the greedy decode: on
\textsc{ParamBench} the cells gain +5.2 to +9.9 EM, largest on
Ministral-8B (from a greedy baseline of 16.7), and for Llama-3.1-8B
the F1 gain reaches +16.8, mostly by repairing structure; on
ComplexFuncBench, where greedy already sits near 29 to 37, gains
stay between +0.8 and +1.7. After PBT all 13 arms improve, by +0.8
to +10.2 EM, and the spread again concentrates on
\textsc{ParamBench}: the seed variants of the same Qwen3-8B cell
span +3.4 to +10.2, the sampling and seed variance of
Appendix~\ref{app:repro}.

\begin{table*}[t]
\centering
\small
\renewcommand{\arraystretch}{1.38}
\setlength{\tabcolsep}{4pt}
\begin{tabular}{llcrrrrrr}
\toprule
& & & \multicolumn{2}{c}{Greedy} & \multicolumn{2}{c}{PGR} & \multicolumn{2}{c}{$\Delta$} \\
\cmidrule(lr){4-5}\cmidrule(lr){6-7}\cmidrule(lr){8-9}
Model & Dataset & Runs & EM & F1 & EM & F1 & EM & F1 \\
\midrule
Qwen3-8B & \textsc{ParamBench} & 16 & 31.74 & 76.11 & 36.95 & 79.47 & \textbf{+5.21} & \textbf{+3.36} \\
Qwen3-14B & \textsc{ParamBench} & 3 & 26.28 & 73.91 & 32.65 & 76.70 & \textbf{+6.37} & \textbf{+2.79} \\
Qwen3-32B & \textsc{ParamBench} & 3 & 31.40 & 77.47 & 37.66 & 80.67 & \textbf{+6.26} & \textbf{+3.20} \\
Ministral-8B & \textsc{ParamBench} & 1 & 16.72 & 38.77 & 26.62 & 65.53 & \textbf{+9.90} & \textbf{+26.76} \\
Llama-3.1-8B & \textsc{ParamBench} & 3 & 9.56 & 47.91 & 15.02 & 64.70 & \textbf{+5.46} & \textbf{+16.79} \\
Qwen3-8B & ComplexFuncBench & 7 & 35.58 & 49.17 & 36.60 & 49.94 & \textbf{+1.02} & \textbf{+0.77} \\
Qwen3-14B & ComplexFuncBench & 1 & 36.62 & 49.99 & 38.31 & 50.56 & \textbf{+1.69} & \textbf{+0.57} \\
Qwen3-32B & ComplexFuncBench & 2 & 36.10 & 49.59 & 36.95 & 50.19 & \textbf{+0.84} & \textbf{+0.59} \\
Llama-3.1-8B & ComplexFuncBench & 2 & 28.96 & 44.15 & 30.26 & 45.41 & \textbf{+1.30} & \textbf{+1.26} \\
\midrule
\multicolumn{3}{l}{Mean over 9 cells} & 28.11 & 56.34 & 32.33 & 62.57 & \textbf{+4.23} & \textbf{+6.23} \\
\bottomrule
\end{tabular}
\caption{Probe-guided reranking applied to instruction-tuned models that
received no fine-tuning, on the 2 datasets of
Figure~\ref{fig:pgrgain}. Every number follows the 5-fold selection
protocol of Section~\ref{subsec:rerank}; repeated runs of a cell share
one greedy decode and are averaged. Gemma-4-12B is excluded by a
prompt-template defect; base variants are excluded because their
greedy decodes in this campaign's pools are frequently unparseable,
turning reranking into format repair rather than selection.}
\label{tab:pgr-bare}
\end{table*}

\begin{table*}[t]
\centering
\small
\renewcommand{\arraystretch}{1.38}
\setlength{\tabcolsep}{3.0pt}
\begin{tabular}{llcrrrrrrl}
\toprule
& & & \multicolumn{2}{c}{Greedy} & \multicolumn{2}{c}{PGR} & \multicolumn{2}{c}{$\Delta$} & \\
\cmidrule(lr){4-5}\cmidrule(lr){6-7}\cmidrule(lr){8-9}
Model & Dataset & Arm & EM & F1 & EM & F1 & EM & F1 & Family \\
\midrule
Qwen3-14B-Base & \textsc{ParamBench} & RS pool & 35.84 & 80.37 & 46.08 & 84.21 & \textbf{+10.24} & \textbf{+3.85} & F-set \\
Qwen3-14B-Base & \textsc{ParamBench} & n12 pool & 35.84 & 80.37 & 44.37 & 82.72 & \textbf{+8.53} & \textbf{+2.35} & F-set \\
Qwen3-14B-Inst & \textsc{ParamBench} & -- & 38.23 & 80.75 & 42.32 & 82.06 & \textbf{+4.10} & \textbf{+1.31} & -- \\
Qwen3-14B-Inst & \textsc{ParamBench} & seed 47 & 33.11 & 79.41 & 38.91 & 81.03 & \textbf{+5.80} & \textbf{+1.61} & F-set \\
Qwen3-8B-Base & \textsc{ParamBench} & -- & 28.67 & 76.36 & 33.11 & 79.82 & \textbf{+4.44} & \textbf{+3.47} & -- \\
Qwen3-8B-Inst & \textsc{ParamBench} & seed 43 & 33.79 & 80.02 & 37.20 & 80.34 & \textbf{+3.41} & \textbf{+0.32} & Cand. \\
Qwen3-8B-Inst & \textsc{ParamBench} & seed 47 & 27.65 & 76.82 & 37.88 & 79.24 & \textbf{+10.24} & \textbf{+2.41} & -- \\
Qwen3-8B-Inst & \textsc{ParamBench} & seed 53 & 25.94 & 76.71 & 31.74 & 79.32 & \textbf{+5.80} & \textbf{+2.61} & F-set \\
Qwen3-8B-Inst & \textsc{ParamBench} & 24-cand.\ pool & 24.57 & 76.96 & 31.06 & 79.53 & \textbf{+6.48} & \textbf{+2.57} & -- \\
Qwen3-8B-Inst & \textsc{ParamBench} & $T=0.95$ & 30.72 & 76.45 & 36.52 & 79.45 & \textbf{+5.80} & \textbf{+3.00} & -- \\
\addlinespace[1.5pt]
Qwen3-14B-Base & ComplexFuncBench & -- & 44.03 & 54.30 & 44.94 & 54.69 & \textbf{+0.91} & \textbf{+0.38} & F-set \\
Qwen3-14B-Inst & ComplexFuncBench & -- & 41.82 & 53.50 & 42.99 & 54.44 & \textbf{+1.17} & \textbf{+0.94} & Cand. \\
Qwen3-8B-Base & ComplexFuncBench & -- & 44.94 & 55.80 & 45.71 & 56.06 & \textbf{+0.78} & \textbf{+0.27} & -- \\
Qwen3-8B-Inst & ComplexFuncBench & -- & 42.34 & 53.92 & 43.12 & 54.11 & \textbf{+0.78} & \textbf{+0.19} & F-set \\
\midrule
\multicolumn{3}{l}{Mean over the 13 arms drawn in Figure~\ref{fig:pgrgain}} & 34.74 & 70.87 & 39.35 & 72.64 & \textbf{+4.61} & \textbf{+1.76} & \\
\bottomrule
\end{tabular}
\caption{Probe-guided reranking applied on top of PBT, on the 2
datasets of Figure~\ref{fig:pgrgain} and under the protocol of
Table~\ref{tab:pgr-bare}. Arms differ in training seed, sampling
temperature, pool size, or archived pool; a dash marks the default
arm. Both archived pools of the Qwen3-14B-Base \textsc{ParamBench}
cell are listed; a dash in the last column marks an arm whose pool
was not retained, so the winning family could not be recomputed.
Greedy values are recomputed from the archived pools.}
\label{tab:pgr-pbt}
\end{table*}

\subsection{Results by Difficulty Level}
\label{app:levels}

The per-level results of Figure~\ref{fig:levels} rest on uneven
sample sizes: the \textsc{ParamBench} test split has 20/72/64/85/52
instances at L1 to L5, and the API-Bank evaluation set has
40/73/54/7/52. The two datasets place their headroom differently.
On \textsc{ParamBench} the seed model already solves L1 at 85 EM
and fails L5 completely, so the gains of PBT+PGR concentrate on the
hard levels, as the +11, +5, and +12 labels of
Figure~\ref{fig:levels} show for L3, L4, and L5. On API-Bank the
first four levels are close to saturated, between 81 and 88 EM, and
the remaining headroom sits at L5, where the gain is +13. Two
reading notes apply. The API-Bank L4 cell rests on only 7 instances
and is best read together with its neighbors, and the L1 cells of
both datasets are small as well, so the flat L1 bars mean saturation
rather than failure of the methods. The middle levels also split: L3
gains 11 points while L2 gains only 3, although both start far from
the ceiling. By the grading rule of Section~\ref{sec:parambench},
an L2 call is at most one level deep and carries no conditional
dependencies, so the structural patterns that the filtered training
data teaches have less room to help there. The overall picture
matches the main text: self-training with a probe filter does not
manufacture ability on levels the seed model never solves alone,
but where a level is within reach, the filtered data and the
reranker convert unstable successes into stable ones.

\paragraph{Released artifact.}
The package contains the material behind this appendix: the
benchmark instances, the 81 frozen API schemas, the probe, PBT, and
PGR implementations of Section~\ref{sec:method}, the per-call
adaptation and evaluation pipeline, and the scripts that regenerate
the figures and appendix tables. The aggregate scores behind
Figures 3 to 5 ship with it, so those figures regenerate without a
GPU, and an offline replay backend runs the evaluation without
network access. A configuration example and the exact dependency
list are included; all entry points are Python modules, with unit
tests covering split construction, adaptation, and training arms.

\end{document}